\documentclass[runningheads]{llncs}
\usepackage{graphicx}

\usepackage{subfig}
\usepackage{mwe}

\usepackage{wrapfig}

\begin{document}
\title{Robust Object Detection with Multi-input Multi-output Faster R-CNN}
%
%

\author{Sebastian Cygert\inst{1}\orcidID{0000-0002-4763-8381} \and \\
Andrzej Czyżewski\inst{1}\orcidID{0000-0001-9159-8658}}
\authorrunning{S. Cygert et al.}
%
\institute{Gdańsk University of Technology \\Faculty of Electronics, Telecommunication and Informatics \\Multimedia Systems Department \\
\email{sebcyg@multimed.org}\\ 
}
\maketitle              
\begin{abstract}
Recent years have seen impressive progress in visual recognition on many benchmarks, however, generalization to the real-world in out-of-distribution setting remains a significant challenge. A state-of-the-art method for robust visual recognition is model ensembling. however, recently it was shown that similarly competitive results could be achieved with a much smaller cost, by using multi-input multi-output architecture (MIMO).  
In this work, a generalization of the MIMO approach is applied to the task of object detection using the general-purpose Faster R-CNN model. It was shown that using the MIMO framework allows building strong feature representation and obtains very competitive accuracy when using just two input/output pairs. Furthermore, it adds just 0.5\% additional model parameters and increases the inference time by 15.9\% when compared to the standard Faster R-CNN. It also works comparably to, or outperforms the Deep Ensemble approach in terms of model accuracy, robustness to out-of-distribution setting, and uncertainty calibration when the same number of predictions is used. This work opens up avenues for applying the MIMO approach in other high-level tasks such as semantic segmentation and depth estimation.

\end{abstract}
\section{Introduction}
Convolutional Neural Networks (CNNs) have recently become a standard method for image processing as they achieve excellent results on many benchmarks. Despite their impressive performance, the current machine learning techniques lack robustness when presented with an image that does not follow the training dataset distribution (out-of-domain data). It was shown that the current models are vulnerable to noisy input \cite{corruptions,fourier}, novel weather conditions \cite{wintercoming}, and background changes \cite{background}, which creates safety considerations for models deployed to the real world, e.g., autonomous driving or medical applications. To improve visual recognition models’ robustness it was proposed to use data augmentations which change objects' appearance, for example by using style-transfer \cite{cnnbiased,towards} data augmentation or color distortions \cite{simclr,gta}.


Additionally, the current models are often overconfident in their predictions \cite{calibration}, and the problem becomes more evident with out-of-domain data \cite{canyoutrust,gta}. 
Sampling based-methods were shown to obtain very good results in terms of accuracy, out-of-domain robustness and improving model predictive uncertainty \cite{medconf,canyoutrust,gta}. The gold standard is the ensemble approach \cite{ensembles_1990}, which involves combining the output of several, diverse models. Several methods were developed to reduce the high computational cost, such as test-time dropout \cite{Gal} or batch ensemble \cite{batchensemble}. Another competitive approach is the m-heads model \cite{mheads}, which can be viewed as an ensemble with parameter sharing in the first layers of the network. Those methods, however, do not always match ensembling accuracy, as the success of sampling-based methods lies in the diversity of the predictions, which is a challenging problem \cite{deepensemble,pitfalls}. 


Recently, the literature on obtaining many predictions from one model using a single inference step has increased. These methods were inspired by the compression methods, which show that it is possible to remove even 90\% of the weights, without affecting the final model accuracy \cite{nipsprune,ticket}. Therefore, instead of compressing the model, it should be possible to fit more than one subnetwork within the main network. For example, \cite{superposition,supermasks} use a single model in the multi-task setting, and the latter approach retrieves a subnetwork (from the main model) to efficiently solve the target task. Another method uses the multi-input multi-output (MIMO) approach, where a single model makes multiple predictions simultaneously. MIMO was shown to only slightly increase the computational cost, while matching the accuracy of model ensembling and was showcased on the image classification task. 
Yet, whether the MIMO approach would work in multi-task setting such as object detection, particularly when regressing objects’ localizations is unknown. In this work, the MIMO method is adapted for object detection tasks and further evaluated. To summarize, the contributions of this work are as follows: 
\begin{itemize}
    \item The multi-input multi-output model was adapted to the object detection task and the architectural changes and implementation details are presented, 
    \item It was shown that such a model acting as a strong regularizer brings significant improvements in terms of in-domain and out-of-domain accuracy, and in classification calibration, by adding only a small computational cost at inference time,
    \item The robustness of the MIMO approach robustness is presented by comparing its results to different Deep Ensemble approaches, which it outperforms (unless a larger number of models are used for the Deep Ensemble) or matches in accuracy.
    
\end{itemize}


\section{Method}


The standard Faster R-CNN model consists of two main modules: first, region proposals are generated using a Region Proposal Network (RPN), and in the second stage, the region proposals are classified and refined using a Region of Interest Pooling network. Computation of the feature map computation is shared between both networks and obtained using some standard CNN architecture. The RPN predicts a set of candidate boxes (anchors), for which it predicts the probability of the anchor being an object and its coordinates. In the final stage, each region proposal is processed by classification layers and regression layers. The first outputs a logit vector $z \in \mathbf{R}^K$ over all $K$ classes, per anchor. The second outputs bounding-box regression offsets, which are used to refine the initial bounding boxes. Finally, a softmax function is applied $ p = softmax(z)$, which results in a list of predicted class probabilities. The whole model is trained using a multi-task loss function, consisting of the classification part (cross-entropy loss) and the regression part of the bounding boxes localizations (L1 smooth loss).

\begin{figure*}[t]
    \centering
    \includegraphics[width=0.99\linewidth,clip]{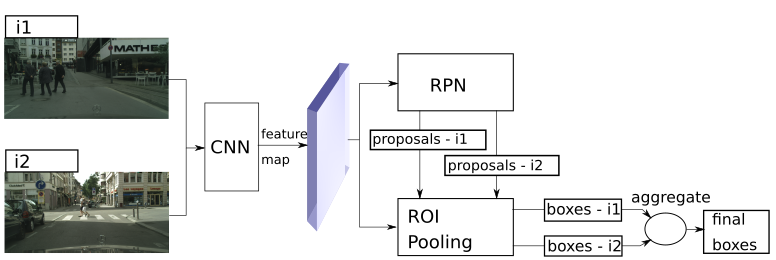}
    \caption{Architecture of the proposed MIMO Faster R-CNN. Both images are sampled independently during training, and each subchannel in the network is responsible for predicting boxes in the corresponding image. During testing, both inputs to the network are the same, and the final results are obtained by running aggregation on both channel results.}
    \label{fig:arch}
\end{figure*}

In this work, the multi-input multi-output approach is applied to the Faster R-CNN model, the architecture overview is presented in the Fig. \ref{fig:arch}. 
To adapt the Faster R-CNN model into the MIMO framework following changes were applied:
\begin{itemize}
    \item Multiply the number of input channels by $M$ (ensemble size),
    \item Region Proposal Network now outputs $M$ sets of region proposals (each per input image),
    \item The ROI Pooling layer independently processes $M$ set of proposals    and the outputs need to be aggregated at test-time. This can be done using the standard non-maximum suppression (NMS) method or more advanced methods, such as Weighted Boxes Fusion \cite{wbf}.
\end{itemize}

Note that the feature map (output from the convolutional backbone) is of the same size as before, however, now it contains information about $M$ images, without forcing any explicit structure on how to share the information from different images. Using that shared feature map RPN returns $M$ independent sets of region proposals, which requires changing the RPN loss function to:

\setlength{\belowdisplayskip}{0pt} \setlength{\belowdisplayshortskip}{0pt}
\setlength{\abovedisplayskip}{0pt} \setlength{\abovedisplayshortskip}{0pt}

\begin{equation}
    L(\{p_i\}, \{t_i\}) = \sum_{m=1}^M (\frac{1}{N_{cls}} \sum_{i}{L_{cls}(\hat{p_{im}}, p_{im})} \\ 
    + \frac{\lambda}{N_{box}} \sum_{i}{p_i smooth_{L1}(\hat{t_{im}} - t_{im})})
\end{equation}

\noindent where $i$ is the anchor index, and $m$ is the index of input/output pair. $\hat{t_{im}}$ are predicted parametrized bounding boxes, $\hat{p_{im}}$ are predicted probabilities of the anchor being an object, and $t_{im}$, $p_{im}$ are the ground-truth counterparts. The equation is normalized by $N_{cls}$ - mini-batch size, and $N_{box}$ - number of anchors. $L_{cls}$ is simply the log loss over two classes (object vs. not object). Note that when $M = 1$, this refers to the standard RPN loss in the Faster R-CNN. Similarly, the loss for the ROI layer, now becomes a sum over $M$ input/output pairs.

During training, each input is being sampled independently. However, during testing, the input is repeated $M$ times so that $M$ possibly different outputs are obtained for \textbf{the same input image}. It was empirically shown in the task of image classification that each of the $M$ outputs provides good accuracy on its own and that the results are diverse enough, which allows them to be efficiently combined. In practice, $M = 2$ is often used. For more complex tasks, the network capacity does not allow processing a larger number of images in parallel. This agrees with the literature on model compression which shows that usually modest compression rates (up to 50\%) are achievable, when using structured pruning (removing whole filters) \cite{nipsprune,structure}.

After $M$ sets of results are obtained, they need to be efficiently combined together. A standard approach in object detection to reduce redundant boxes is the NMS algorithm, which clusters together detections with a high overlap, and keeps only detections with a high confidence. Such a procedure might be non-optimal when combining predictions from different models. Recently, the Weighted Boxes Fusion (WBF) \cite{wbf} method was proposed. It efficiently combines different predictions by updating the final bounding box coordinates by using the confidence-weighted average of coordinates forming a cluster. Similarly, the final confidence score is also an average of all boxes forming a cluster. In this work, both aggregation methods (NMS and WBF) are evaluated.

\section{Experiments}

\subsection{Experimental setting}

\textbf{Datasets}. For the evaluation, Cityscapes \cite{cityscapes}, Berkeley Deep Drive (BDD) \cite{bdd} and MS-COCO \cite{coco} datasets were used. BDD dataset  allows to study the model robustness to natural distributional shift when the model is trained on daytime images and evaluated on nighttime images, the same as in \cite{wintercoming}. To further test the robustness of the trained models a corrupted version \cite{corruptions} of the Cityscapes dataset was used. It included a number of synthetically generated distortions types grouped into four main categories: noise, blur, weather corruptions and digital noise. Each corruption has five levels of intensity, and for simplicity the distortions were applied using a medium intensity level. Usage of this benchmark is a popular way to measure models’ robustness in the o.o.d. setting \cite{towards,fourier,mimo,mixmo,wintercoming,corruptions,TODOICML,canyoutrust}.


\textbf{Implementation details}. In our experiments, the MMDetection framework \cite{mmdetection} was used. For the Cityscapes experiments, the models were trained for 64 epochs, using SGD optimizer, with an initial learning rate of 0.01 and a learning rate step reduction by a factor of 10 at epoch 48, similar as in \cite{wintercoming}. All models were trained on a single GPU (Tesla V100) using a batch size of 6. During the training standard vision-based augmentations are applied: horizontal flipping and random resize. All of the models were pretrained on ImageNet \cite{imagenet}, as it is a standard in the community. For BDD and COCO datasets, the training lasted for 12 epochs, with learning rate reductions at epochs 8 and 11. Results are reported on the held-out validation sets.

The color jittering data augmentation was applied using the Albumentations library \cite{albumentations} with default parameters and the following transformations: random changes in brightness, contrast, saturation and hue. In addition, style-transfer data augmentation was also used for some models, to improve the diversity of the ensemble approach.
A standard procedure was applied, as in \cite{cnnbiased,wintercoming}, for style-transfer: as the source of the style images, Kaggle’s Painter By Numbers dataset was used, and during training, a stylized image was sampled with probability $p = 0.5$, otherwise, the original image was used.

\textbf{Uncertainty estimation}. The Expected Calibration Error (ECE) is one of the ways to compute model calibration \cite{calibration}. Based on the confidence of the prediction, they are partitioned into \textit{M} bins, and the ECE is computed as the weighted average of the calibration error within each bin:

\begin{equation}
ECE(c) = \sum_{m=1}^M \frac{|B_m|}{n}|acc(B_m) - conf(B_m)|
\end{equation}

\noindent with $B_m$ being the set of indices of the samples for which the prediction confidence falls into the $m$th interval. A lower score means better calibration. It is also possible to compute a calibration score for the regression of the bounding box localization; however this is beyond the scope of our work.

\subsection{MIMO Faster R-CNN}
\setlength{\intextsep}{-14pt}%
\setlength{\columnsep}{6pt}%
\begin{wrapfigure}{r}{0.5\textwidth}
  \centering
  \includegraphics[width=0.5\textwidth]{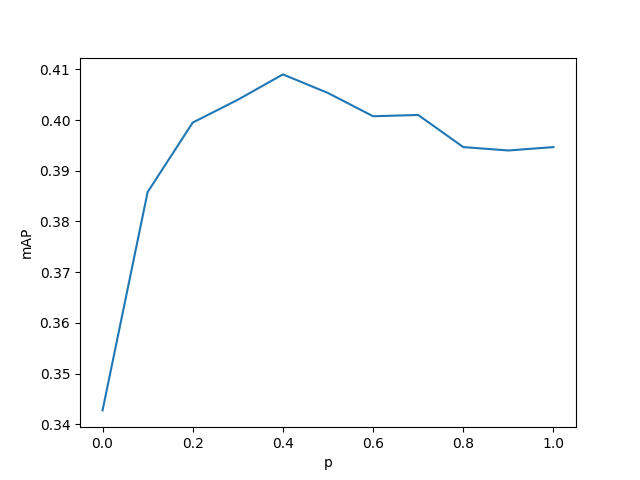}
  \caption{Model accuracy (mAP) on Cityscapes dataset as a function of probability $p$ that the same images are sampled, when the model is trained with $M=2$ input/output pairs. 
  \label{fig:patchsize}
  }
\end{wrapfigure}


Standard MIMO architecture struggles with fitting more subnetworks, especially on more challenging datasets. For example, in \cite{mimo} the authors found that when training a ResNet-50\cite{resnet} classifier on the ImageNet dataset with $M = 2$, the model performed worse than a baseline. It was hypothesized that this happens when the main network does not have sufficient capacity to correctly classify two independent images at once. To improve that, the authors proposed relaxing independence between the inputs and added another hyperparameter $p$, which defines the probability that the networks use the same data during training. Namely when $p = 0$, both images are sampled independently, and when $p = 1$, the training images are the same. As a result, in our first experiments on the Cityscapes dataset, a model with $M = 2$ input/output pairs was trained, and the $p$ parameter was varied to see how it affected the final model performance (Fig. \ref{fig:patchsize}). 

At $p = 0$ the inputs were fully independent, however the final performance is limited by the network capacity. As the $p$ grew, the subnetworks used the same image during training (with $p$ probability), which allowed some of the features to be shared, which improved the performance. The performance peaked at $p = 0.4$ and then is slightly decreased. It is similar result to that described in \cite{mimo}, when using ResNet-50 for ImageNet classification task. As a result, further experiments were performed using $M = 2$, and $p = 0.4$. 

Further, the results are compared with the standard Faster R-CNN model and Deep Ensemble approach (also consisting of $M = 2$ models) (Table \ref{tab:baselines}). 
First, the MIMO Faster R-CNN outperformed a single model, improving the mAP score from 0.386 to 0.409. It also slightly outperformed the Deep Ensemble model. Importantly, the MIMO model brought only a slight increase in the parameters compared to the standard Faster R-CNN (from 41.38M to 41.4M). In terms of inference time (as measured on a Tesla V-100 GPU), it has increased by 15.9\% (from 88ms to 102 ms per image). Note, that when applying the MIMO framework to the image classification task the increase in inference time was very small (around 1\%) \cite{mimo}. For the task of object detection, a larger increase in the processing time is attributed to \textit{M}-times larger number of proposal regions being processed and the additional aggregation method. However, the processing time was still significantly smaller when compared to the Deep Ensemble method.

It is important to note that starting the training with ImageNet weights was crucial for the Cityscapes dataset (for all models). Additionally, when training the MIMO Faster R-CNN, one must also copy the ImageNet to the new filters (in the first channel). For the Deep Ensemble approach, the WBF aggregation method provided better results, yielding an improvement in the mAP score of 0.01, over the NMS approach. Overall, the WBF method performed the same or better than the NMS method, and all of the results were achieved using WBF aggregation.
We also conducted experiments on the naive m-heads architecture \cite{mheads}, in which the backbone was shared and the RPN and ROI nets were doubled. However, such an approach resulted in poor performance (mAP score of 0.378) and larger increase of number of parameters (55.9M). Such poor performance might be a result of the non-optimal structure of the proposed m-heads approach, and as such other variants could be explored.

\begin{table}[t]
\centering
\caption{Accuracy and computational cost of different methods.}
\label{tab:baselines}
\begin{tabular}{|l|l|l|l|}
\hline
Model & mAP & num. params. & inf. time \\
\hline
Baseline & 0.386 & 41.384M & 0.088\\
\hline
MIMO (M=2) & 0.409 & 41.397M & 0.102\\
\hline
Deep Ensemble (M=2) & 0.406 & 82.768M & 0.176\\
\hline
\end{tabular}
\end{table}

\subsection{Robustness and uncertainty}
In this section, further experiments are described which focus on robustness and uncertainty estimation. First, it can be noted that the accuracy in the o.o.d. setting is severely impacted (Table \ref{tab:robustness}), as it was previously shown in the literature. For the baseline model the accuracy on the corrupted version of the dataset was equal to 0.106. The accuracy was significantly improved when using the MIMO approach (0.172), outperforming Deep Ensemble by a large margin (0.116).

Further, the impact of adding color jittering data augmentation was measured. As expected, it improved the accuracy of the baseline model in the o.o.d. setting. On the other hand, the MIMO approach, did not result in significant changes, except for slightly improving model calibration. Deep Ensemble also benefited from the added data augmentation, but it clearly lacks the robustness of the MIMO approach (e.g., 0.134 mAP score in the o.o.d. testing, compared to the 0.172 of the MIMO approach). When using color jittering, no significant changes were observed when measuring the impact of the $p$ value on the final accuracy (as in the Fig. \ref{fig:patchsize}), and as a result $p=0.4$ was further used.

It can also be noticed that the MIMO approach provided the best classification calibration results above of the tested models. The ECE score on the clean dataset equalled 0.045 (compared to 0.066 of the baseline model), and was further reduced to 0.042 when color jitter data augmentation was used. The ECE in the o.o.d. setting was again the lowest out of the evaluated methods, also significantly outperforming the Deep Ensemble approach. 

\begin{table}[t]
\centering
\caption{Models' accuracy and calibration using different models and augmentation methods. Last two columns present results for corrupted Cityscapes. CJ stands for the color jittering augmentation and DE for the Deep Ensemble.}
\label{tab:robustness}
\begin{tabular}{|l|l|l|l|l|}
\hline
Model & mAP & ECE & c-mAP & c-ECE\\
\hline
Baseline & 0.386 & 0.066 & 0.106 & 0.113\\
\hline
CJ & 0.388 & 0.064 & 0.124 & 0.115\\
\hline
MIMO ($M=2$) & \textbf{0.409} & 0.045 & \textbf{0.172} & 0.075\\
\hline
MIMO ($M=2$) + CJ & 0.408 & \textbf{0.04} & \textbf{0.172} & \textbf{0.071}\\
\hline
DE ($M=2$, baseline) & 0.406 & 0.068 & 0.116 & 0.124\\
\hline
DE ($M=2$, CJ) & 0.408 & 0.062 & 0.134 & 0.112\\
\hline
\hline
DE (MIMO, \textit{M=2}) & \textbf{0.426} & 0.05 & 0.184 & 0.087\\
\hline
DE (MIMO+CJ, \textit{M=2}) & 0.425 & \textbf{0.046} & \textbf{0.186} & \textbf{0.068}\\
\hline
DE (baseline, \textit{M=5}) & 0.417 & 0.078 & 0.122 & 0.129\\
\hline
DE (CJ + style, \textit{M=5}) & 0.421 & 0.075 & 0.139 & 0.114\\
\hline

\end{tabular}
\end{table}






A potential critique of the evaluated Deep Ensemble approach is that a very small ensemble size was used, and that the models' diversity is limited. As such, the ensemble method was also tested when 5 models were used, which was shown in the literature to provide good results already \cite{canyoutrust}. Additionally, one of the ensembles consisted of models of which some used color jittering data augmentation, and some used style-transfer data augmentation, to improve ensemble diversity. That setting allowed the Deep Ensemble approach to obtain very competitive results (Table \ref{tab:robustness}, bottom part). The mAP increased to 0.421 and 0.139 on the clean and corrupted versions of the Cityscapes datasets, respectively (note that the accuracy in the o.o.d. is still worse than when using the MIMO approach). 
It was also checked whether using a MIMO Faster R-CNN models ensemble could further improve the results. When combining the outputs from two MIMO Faster R-CNN models, an impressive mAP of 0.426 was obtained on the clean dataset, and 0.184 on the corrupted version, which is an improvement over the Deep Ensemble approach consisting of 5 models. The usage of color jittering has improved the model calibration. Overall, these results further confirm the robustness of the MIMO Faster R-CNN model. Sample detection are presented in the Fig. \ref{fig:qual1}.

It is also interesting to look at the accuracy of the MIMO Faster R-CNN when using only one output. In such a scenario, model accuracy equals 0.405, which is a 0.004 drop compared to the full MIMO approach, but it is still a significant improvement over the baseline (0.386). This shows that the MIMO framework acts as a strong regularizer during training, which leads to strong feature representations.

\begin{table}[t]
\centering
\caption{Accuracy on BDD Dataset when training on daytime images. $M=2$ was used.}
\label{tab:other}
\begin{tabular}{|l|l|l|}
\hline
Model & BDD-day & BDD-night \\
\hline
Baseline & 0.293 & 0.233\\
\hline
CJ & 0.293 & 0.237\\
\hline
MIMO $(p=0.7)$ & 0.301 & 0.244\\
\hline
MIMO + CJ $(p=0.7)$ & 0.3 & 0.241\\
\hline
DE (baseline) & 0.3 & 0.24 \\
\hline
DE (CJ) & \textbf{0.302} & \textbf{0.246} \\
\hline
\end{tabular}
\end{table}

The proposed method was also evaluated on the \textbf{BDD dataset} (Table \ref{tab:other}). The model was trained using daytime images only and evaluated on daytime and nighttime images in this setting. Overall, compared to the standard training, the MIMO approach improved the accuracy on the clean daytime images from 0.293 to 0.301 and on nighttime images (o.o.d. test) from 0.233 to 0.244. The probability $p$ of sampling the same images during training also had to be further increased to observe improvements when using the MIMO model.
This might be because a BDD is more challenging and includes a larger and more diverse set of images than the Cityscapes dataset. In that setting, the results of the MIMO approach are very similar to those obtained by Deep Ensemble. Looking at the single model accuracy within the MIMO method, it was found that it achieved almost the same accuracy (0.3 mAP value) as the full model. However, since the BDD dataset is very challenging, and the probability $p$ of sampling the same images had to be increased, the outputs from single channels are no longer diverse, limiting the accuracy of the MIMO framework for this dataset.
We experimented with a larger backbone (ResNet-101), but it provided a similar increase over the standard model. 

\setlength{\intextsep}{15pt}%
\setlength{\columnsep}{5pt}%

\begin{wrapfigure}{r}{0.5\textwidth}

  \centering
  \includegraphics[width=0.47\textwidth]{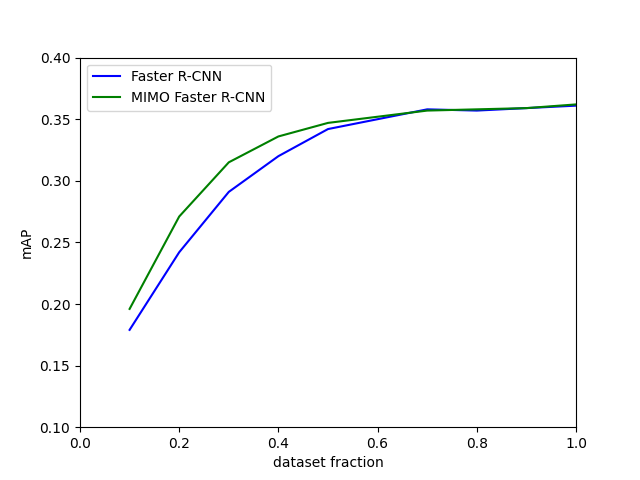}
  \caption{Accuracy of the standard and MIMO-based Faster R-CNN  on the COCO dataset when using only a fraction of the training dataset. 
  \label{fig:coco}
  }
\end{wrapfigure}

When evaluating the model on the COCO dataset no gains in accuracy were observed. Given the hypothesis that significant gains of the MIMO approach come from the regularization property, it should work better when using only the fraction of the training dataset. In fact, such an observation was made and it was shown that the MIMO approach was in particular useful in the low data regime (Fig. \ref{fig:coco}). Each model was trained for the same number of steps. MIMO framework was in particular effective when less then 50\% of training dataset was used, e.g. when using 30\% of the data, using the MIMO approach improved the accuracy from 0.291 to 0.315 of the mAP score.

\subsection{Discussion}

\newlength{\tempheight}
\newlength{\tempwidth}

\newcommand{\rowname}[1]
{\rotatebox{90}{\makebox[\tempheight][c]{\textbf{#1}}}}
\newcommand{\columnname}[1]
{\makebox[\tempwidth][c]{\textbf{#1}}}

\begin{figure*}[t]
\setlength{\tempwidth}{.32\linewidth}
\settoheight{\tempheight}{\includegraphics[width=\tempwidth]{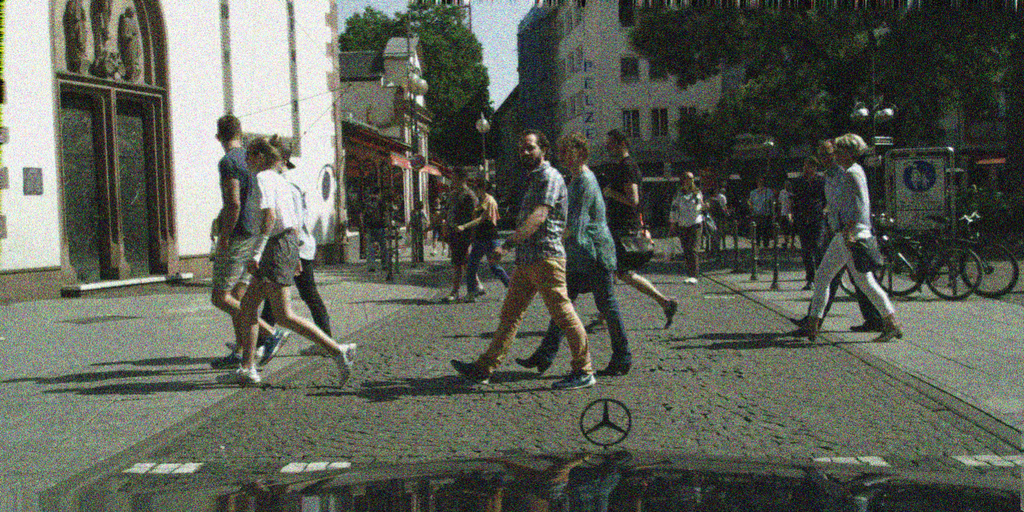}}%
\centering
\hspace{\baselineskip}
\\
\rowname{Baseline}
\subfloat{\includegraphics[width=\tempwidth]{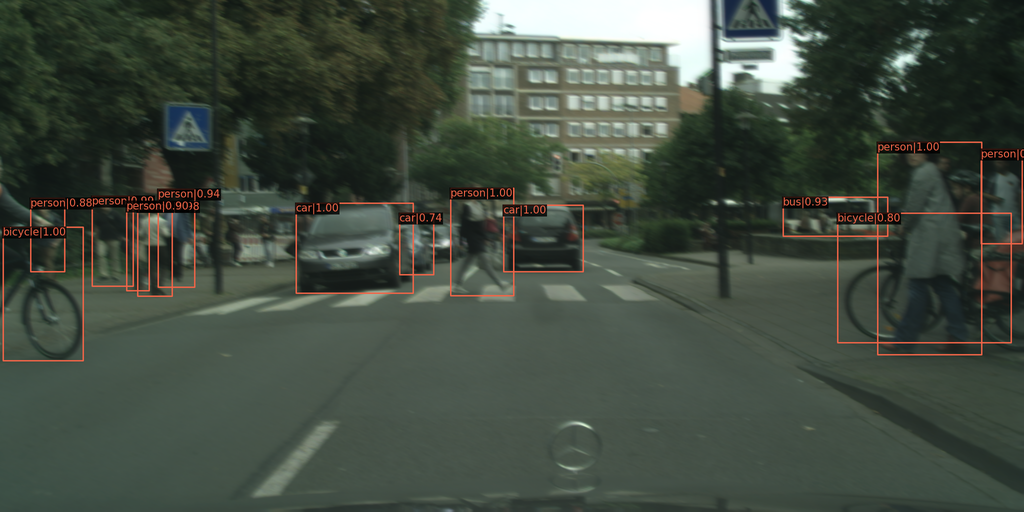}}\hfil
\subfloat{\includegraphics[width=\tempwidth]{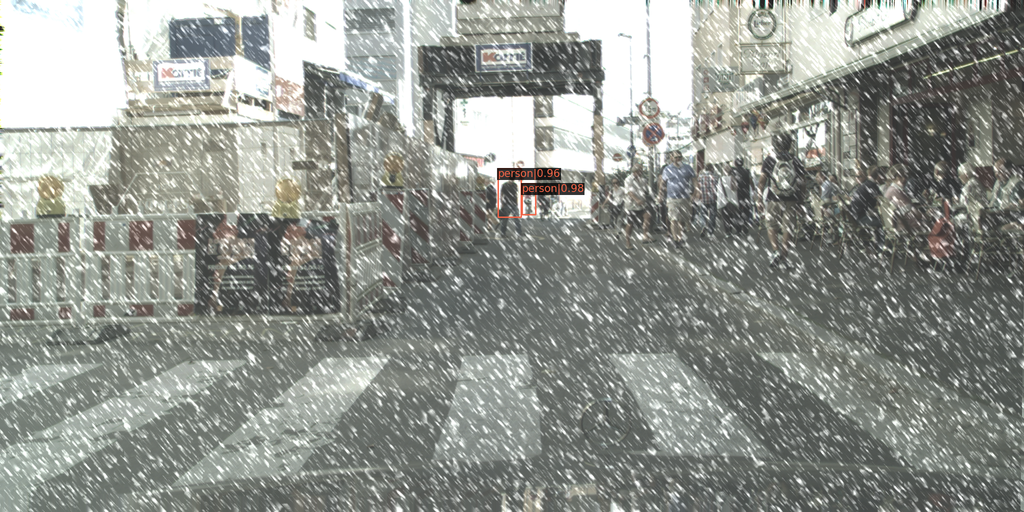}}\hfil
\subfloat{\includegraphics[width=\tempwidth]{imgs/fig_4_1.png}}\hfil
\rowname{MIMO}
\subfloat{\includegraphics[width=\tempwidth]{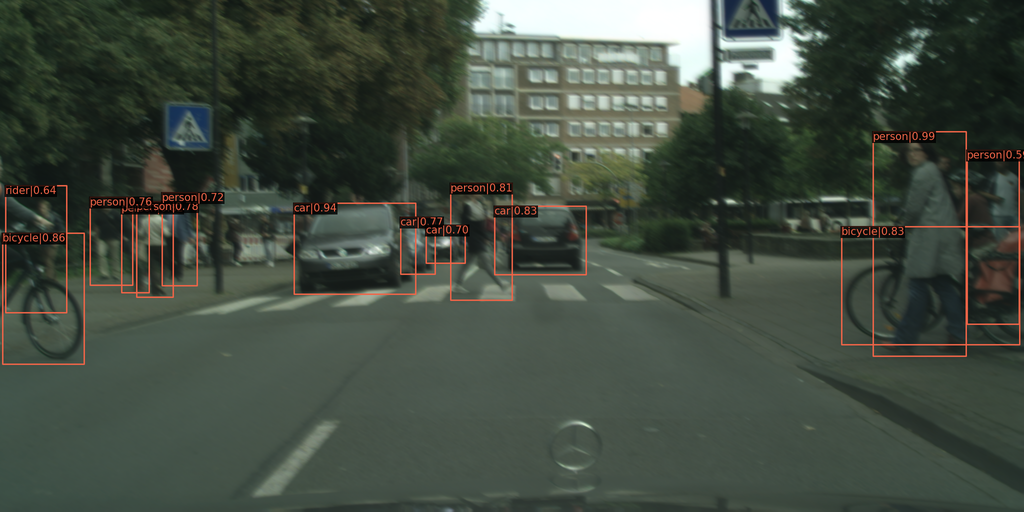}}\hfil
\subfloat{\includegraphics[width=\tempwidth]{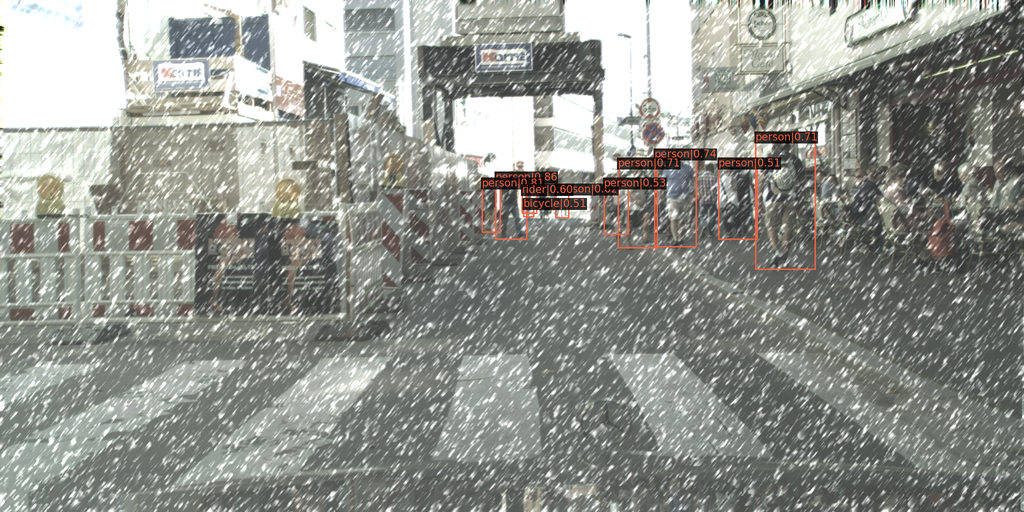}}\hfil
\subfloat{\includegraphics[width=\tempwidth]{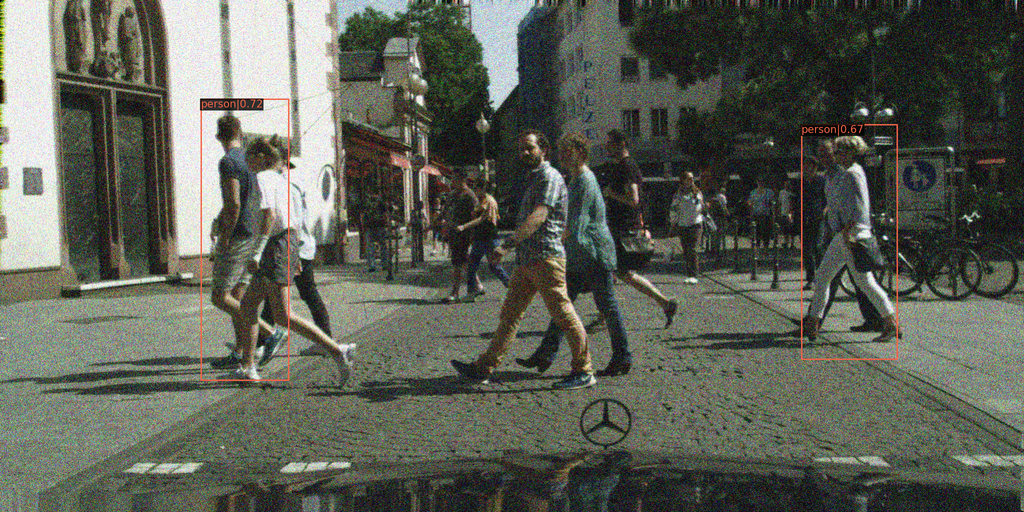}}\hfil
\caption{Detection results for the baseline and MIMO Faster R-CNN on different distortion types (motion blur, snow effect, Gaussian noise in the consecutive columns). Note, smaller confidence values for the MIMO model (i.e., 1st column). MIMO model performs on par or better than the standard model, however the corruptions vulnerability remains challenging (3rd column). Best viewed in digital format.}
\label{fig:qual1}
\end{figure*}

The experiments showed that the MIMO approach could bring significant accuracy improvements compared to the standard training, when using just $M = 2$ input/output pairs. Further, using just one output from the MIMO approach brings a significant gain in the model accuracy. The MIMO approach turned out to be especially effective when using a fraction of the original dataset on the COCO dataset. Given those observations, we conjecture that training the model in multi-input multi-output works as a very strong regularizer, which allows the model to build a robust feature representation, and therefore even when using a small \textit{M}, the model can work very well. This finding adds more context to the original MIMO paper \cite{mimo}, which attributed its great performance mainly to ensembling diverse predictions. A similar observation was made in the literature for structured pruning, which showed that model compression could actually increase its performance (at modest compression rates) \cite{structure}.

It was also interesting to note that using specific data augmentation (for example, style transfer, color jittering) was not essential for the MIMO model to significantly improve the accuracy in the out-of-distribution setting. Again, this might indicate that using such texture-invariant data augmentation is unnecessary for the model to increase its robustness when its build representation is strongly regularized. This can be viewed as a complementary finding to a recent work \cite{TODOICML}, which shows that the increased shape bias (using the aforementioned data augmentations) does not necessarily improve model robustness.

A potential drawback of the MIMO approach is that when the task or dataset is especially challenging, it requires increasing the probability $p$ of sampling the same images during training. This reduces the diversity between model outputs and diminishes the potential gains of having multiple outputs. 

The presented results could be further improved. For example, it was shown that using batch repetition during training for the MIMO framework has improved the results \cite{mimo,mixmo}, however, this came at the cost of significantly increased training time. Also, no specific optimization of the hyperparameters for MIMO was performed.

\section{Conclusions}

This work showed that using a multi-input multi-output approach can be generalized to object detection on real-world datasets. 
The MIMO Faster R-CNN model presented very competitive results in terms of model accuracy, uncertainty calibration and-out-of distribution robustness, when using only $M=2$ input/output pairs. The model adds only 0.5\% of model parameters and increases the inference time by 15.9\%. Similar accuracy can also be obtained when using the Deep Ensemble approach, but on the Cityscapes dataset, it required using a larger number of models (and a significantly higher computational cost). The MIMO approach works as a regularizer during training, which significantly increases the accuracy of a single subnetwork compared to the standard training. A current limitation of the MIMO framework is that when the target dataset or task is challenging, the probability $p$ of sampling the same image during training must be increased, limiting the diversity of the MIMO outputs. Some optimizations to the MIMO framework would be an interesting future work on scaling that approach to such a setting. 

The authors believe that this work opens up many directions for further research, including applying the approach to other high-level tasks such as semantic segmentation, and depth estimation, and to other domains of machine learning.

%
%
%
\bibliographystyle{splncs04}
\bibliography{biblio}

\begin{thebibliography}{10}
\providecommand{\url}[1]{\texttt{#1}}
\providecommand{\urlprefix}{URL }
\providecommand{\doi}[1]{https://doi.org/#1}

\bibitem{pitfalls}
Ashukha, A., Lyzhov, A., Molchanov, D., Vetrov, D.P.: Pitfalls of in-domain
  uncertainty estimation and ensembling in deep learning. In: 8th International
  Conference on Learning Representations, {ICLR} (2020)

\bibitem{albumentations}
Buslaev, A., Iglovikov, V.I., Khvedchenya, E., Parinov, A., Druzhinin, M.,
  Kalinin, A.A.: Albumentations: Fast and flexible image augmentations.
  Information  \textbf{11}(2) (2020)

\bibitem{mmdetection}
Chen, K., et~al.: {MMDetection}: Open mmlab detection toolbox and benchmark.
  preprint arXiv:1906.07155  (2019)

\bibitem{simclr}
Chen, T., Kornblith, S., Norouzi, M., Hinton, G.E.: A simple framework for
  contrastive learning of visual representations. In: Proceedings of the 37th
  International Conference on Machine Learning, {ICML} (2020)

\bibitem{superposition}
Cheung, B., Terekhov, A., Chen, Y., Agrawal, P., Olshausen, B.A.: Superposition
  of many models into one. In: Annual Conference on Neural Information
  Processing Systems, {NeurIPS} 2019

\bibitem{cityscapes}
Cordts, M., et~al.: The cityscapes dataset for semantic urban scene
  understanding. In: Proceedings of the IEEE conference on computer vision and
  pattern recognition (2016)

\bibitem{towards}
{Cygert}, S., {Czyżewski}, A.: Toward robust pedestrian detection with data
  augmentation. IEEE Access  \textbf{8} (2020)

\bibitem{gta}
Cygert, S., Wróblewski, B., Woźniak, K., Słowiński, R., Czyżewski, A.:
  Closer look at the uncertainty estimation in semantic segmentation under
  distributional shift. In: 2021 International Joint Conference on Neural
  Networks (IJCNN) (2021)

\bibitem{ticket}
Frankle, J., Carbin, M.: The lottery ticket hypothesis: Finding sparse,
  trainable neural networks. In: 7th International Conference on Learning
  Representations, {ICLR} 2019

\bibitem{Gal}
Gal, Y., Ghahramani, Z.: Dropout as a bayesian approximation: Representing
  model uncertainty in deep learning. In: Proceedings of the 33nd International
  Conference on Machine Learning, ICML. vol.~48 (2016)

\bibitem{cnnbiased}
Geirhos, R., et~al.: Imagenet-trained cnns are biased towards texture;
  increasing shape bias improves accuracy and robustness. In: 7th International
  Conference on Learning Representations, ICLR (2019)

\bibitem{calibration}
Guo, C., Pleiss, G., Sun, Y., Weinberger, K.Q.: On calibration of modern neural
  networks. In: Proceedings of the 34th International Conference on Machine
  Learning-Volume 70 (2017)

\bibitem{nipsprune}
Han, S., Pool, J., Tran, J., Dally, W.: Learning both weights and connections
  for efficient neural network. In: Advances in Neural Information Processing
  Systems (2015)

\bibitem{ensembles_1990}
{Hansen}, L.K., {Salamon}, P.: Neural network ensembles. IEEE Transactions on
  Pattern Analysis and Machine Intelligence  (1990)

\bibitem{mimo}
Havasi, M., et~al.: Training independent subnetworks for robust prediction. In:
  International Conference on Learning Representations, ICLR (2021)

\bibitem{resnet}
He, K., Zhang, X., Ren, S., Sun, J.: Deep residual learning for image
  recognition. In: 2016 {IEEE} Conference on Computer Vision and Pattern
  Recognition. \doi{10.1109/CVPR.2016.90}

\bibitem{corruptions}
Hendrycks, D., Dietterich, T.G.: Benchmarking neural network robustness to
  common corruptions and perturbations. In: 7th International Conference on
  Learning Representations, {ICLR} 2019

\bibitem{deepensemble}
Lakshminarayanan, B., Pritzel, A., Blundell, C.: Simple and scalable predictive
  uncertainty estimation using deep ensembles. In: Advances in Neural
  Information Processing Systems 30 (NeurIPS) (2017)

\bibitem{mheads}
Lee, S., Purushwalkam, S., Cogswell, M., Crandall, D.J., Batra, D.: Why {M}
  heads are better than one: Training a diverse ensemble of deep networks.
  preprint arXiv:1511.06314  (2015)

\bibitem{coco}
Lin, T.Y., et~al.: Microsoft coco: Common objects in context. In: Computer
  Vision -- ECCV 2014

\bibitem{medconf}
Mehrtash, A., Wells, W.M., Tempany, C.M., Abolmaesumi, P., Kapur, T.:
  Confidence calibration and predictive uncertainty estimation for deep medical
  image segmentation. IEEE Transactions on Medical Imaging  (2020)

\bibitem{wintercoming}
Michaelis, C., Mitzkus, B., Geirhos, R., Rusak, E., Bringmann, O., Ecker, A.S.,
  Bethge, M., Brendel, W.: Benchmarking robustness in object detection:
  Autonomous driving when winter is coming. In: Machine Learning for Autonomous
  Driving Workshop, NeurIPS (2019)

\bibitem{TODOICML}
Mummadi, C.K., Subramaniam, R., Hutmacher, R., Vitay, J., Fischer, V., Metzen,
  J.H.: Does enhanced shape bias improve neural network robustness to common
  corruptions? In: Proceedings of the 38th International Conference on Machine
  Learning, {ICML} (2021)

\bibitem{canyoutrust}
Ovadia, Y., et~al.: Can you trust your model's uncertainty? evaluating
  predictive uncertainty under dataset shift. In: Advances in Neural
  Information Processing Systems (2019)

\bibitem{mixmo}
Ram{\'{e}}, A., Sun, R., Cord, M.: Mixmo: Mixing multiple inputs for multiple
  outputs via deep subnetworks. preprint arXiv:2103.06132

\bibitem{imagenet}
Russakovsky, O., et~al.: {ImageNet Large Scale Visual Recognition Challenge}.
  International Journal of Computer Vision (IJCV)  (2015)

\bibitem{wbf}
Solovyev, R.A., Wang, W., Gabruseva, T.: Weighted boxes fusion: Ensembling
  boxes from different object detection models. Image Vis. Comput.  (2021)

\bibitem{structure}
Wen, W., Wu, C., Wang, Y., Chen, Y., Li, H.: Learning structured sparsity in
  deep neural networks. In: Advances in Neural Information Processing Systems
  29 (2016)

\bibitem{batchensemble}
Wen, Y., Tran, D., Ba, J.: Batchensemble: an alternative approach to efficient
  ensemble and lifelong learning. In: 8th International Conference on Learning
  Representations, {ICLR} (2020)

\bibitem{supermasks}
Wortsman, M., Ramanujan, V., Liu, R., Kembhavi, A., Rastegari, M., Yosinski,
  J., Farhadi, A.: Supermasks in superposition. In: Annual Conference on Neural
  Information Processing Systems 2020, NeurIPS 2020 (2020)

\bibitem{background}
Xiao, K.Y., Engstrom, L., Ilyas, A., Madry, A.: Noise or signal: The role of
  image backgrounds in object recognition. In: 9th International Conference on
  Learning Representations, {ICLR} 2021 (2021)

\bibitem{fourier}
Yin, D., Lopes, R.G., Shlens, J., Cubuk, E.D., Gilmer, J.: A fourier
  perspective on model robustness in computer vision. In: Advances in Neural
  Information Processing Systems (2019)

\bibitem{bdd}
Yu, F., et~al.: {BDD100K:} {A} diverse driving dataset for heterogeneous
  multitask learning. In: 2020 {IEEE/CVF} Conference on Computer Vision and
  Pattern Recognition. {IEEE} (2020)

\end{thebibliography}
\end{document}